\documentclass{article}

\usepackage{microtype}
\usepackage{graphicx}
\usepackage{subfigure}
\usepackage{booktabs} % for professional tables
\usepackage{lipsum}  
\setlipsum{%
  par-before = \begingroup\color{gray},
  par-after = \endgroup
}
\usepackage{hyperref}
\usepackage{kotex}

\usepackage[accepted]{icml2024}

\usepackage{amsmath}
\usepackage{amssymb}
\usepackage{mathtools}
\usepackage{amsthm}

\usepackage{pifont}
\usepackage{booktabs} % For formal tables

\usepackage{xcolor}
\usepackage{hyperref}
\usepackage[capitalize,noabbrev]{cleveref}

\theoremstyle{plain}

\theoremstyle{definition}

\theoremstyle{remark}

\newcommand{\eg}{\textit{e}.\textit{g}.}
\newcommand{\Sref}[1]{Section \ref{#1}}

\newcommand{\fref}[1]{Figure~\ref{#1}}

\newcommand{\tref}[1]{Table~\ref{#1}}
\newcommand{\sref}[1]{~\Sref{#1}}
\newcommand{\aref}[1]{Appendix \ref{#1}}
\newcommand{\dref}[1]{\ref{#1}}

\def\swappingselfattention{swapping self-attention}

\def\crossstyleattention{cross style-attention}

\def\ours{visual style prompting}
\def\Ours{Visual style prompting}

\definecolor{darkgreen}{rgb}{0.032, 0.6392, 0.2039}
\icmltitlerunning{Visual Style Prompting with Swapping Self-Attention}

\begin{document}

\twocolumn[
\icmltitle{
Visual Style Prompting with Swapping Self-Attention
%Swapping Self-attention in Diffusion Models for Style Try-On
% Implicit Decoupling of Style\&Content via Selective Cross-Style Attention in DMs
}



\icmlsetsymbol{equal}{*}

\begin{icmlauthorlist}
\icmlauthor{Jaeseok Jeong}{equal,yyy,comp}
\icmlauthor{Junho Kim}{equal,comp}
\icmlauthor{Yunjey Choi}{comp}
\icmlauthor{Gayoung Lee}{comp}
\icmlauthor{Youngjung Uh}{yyy}
\end{icmlauthorlist}

\icmlaffiliation{yyy}{Yonsei University}
\icmlaffiliation{comp}{NAVER AI Lab}

\icmlcorrespondingauthor{Youngjung Uh}{yj.uh@yonsei.ac.kr}

\icmlkeywords{Machine Learning, ICML}

\vskip 0.3in
]



\printAffiliationsAndNotice{\icmlEqualContribution} % otherwise use the standard text.

% 내용 시작
\begin{abstract}

In the evolving domain of text-to-image generation, diffusion models have emerged as powerful tools in content creation.
Despite their remarkable capability, existing models still face challenges in achieving controlled generation with a consistent style, requiring costly fine-tuning or often inadequately transferring the visual elements due to content leakage. To address these challenges, we propose a novel approach, \ours, to produce a diverse range of images while maintaining specific style elements and nuances. During the denoising process, we keep the query from original features while swapping the key and value with those from reference features in the late self-attention layers. This approach allows for the visual style prompting without any fine-tuning, ensuring that generated images maintain a faithful style. Through extensive evaluation across various styles and text prompts, our method demonstrates superiority over existing approaches, best reflecting the style of the references and ensuring that resulting images match the text prompts most accurately. Our project page is available \href{https://curryjung.github.io/VisualStylePrompt/}{here}.

\end{abstract}
\section{Introduction}
\label{introduction}

\begin{figure}[t]
\vskip 0.2in
\begin{center}
\centerline{\includegraphics[width=0.9\columnwidth]{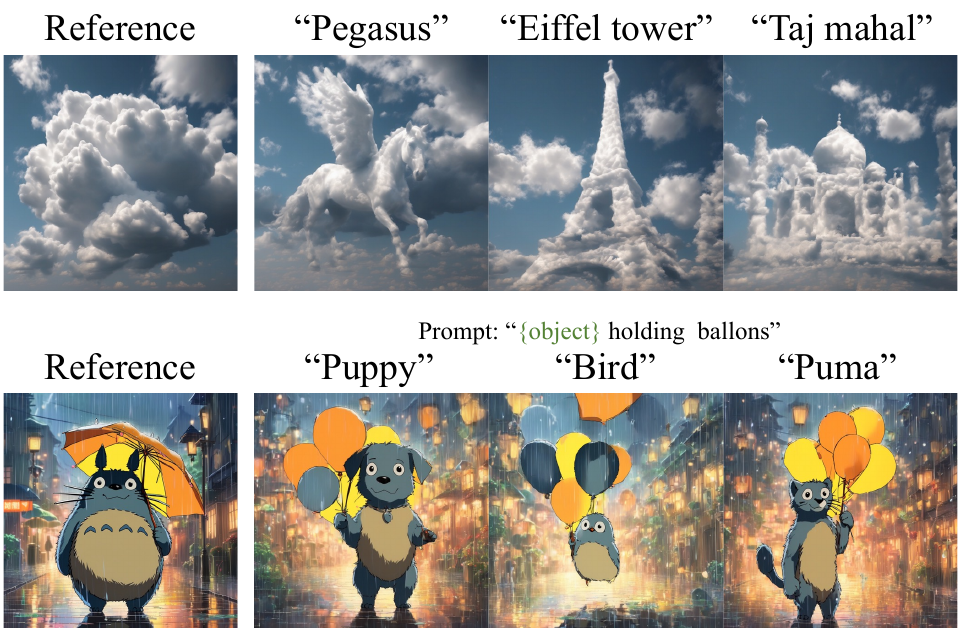}}
\caption{
We tackle visual style prompting, reflecting
style elements from reference images and 
contents from text prompts,
in a training-free manner.}
\label{fig:teasor_v0}
\end{center}
\vskip -0.3in
\end{figure}

\begin{figure}[t]
\vskip 0.2in
\begin{center}
\centerline{\includegraphics[width=\columnwidth]{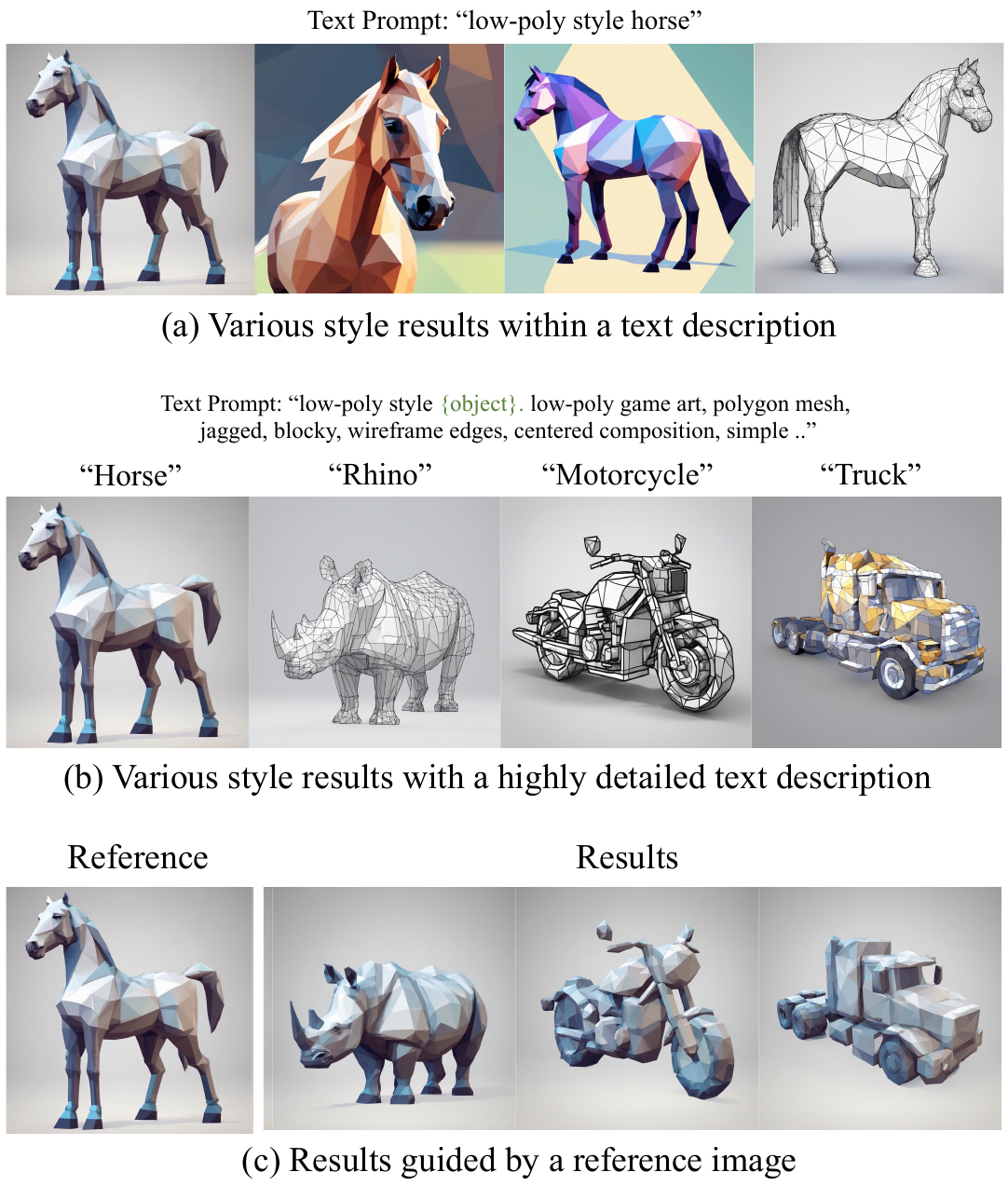}}
\caption{\textbf{Ambiguity of text prompts.} (a) Ambiguity of text leads to different results within the same style description. (b) Even a detailed style description does not guarantee the generation of the same style images since it has many variants that can hardly be constrained using only text prompts. (c) Reference images can specify detailed visual elements. }
\label{fig:text_prompt_not_enough}
\end{center}
\vskip -0.3in
\end{figure}

Text-to-image diffusion models (T2I DMs) have excelled at synthesizing stunning images that correspond to given text prompts, essentially enabling content creation \cite{rombach2022high,ramesh2021zero}. Despite their remarkable capability, inherent discrepancies between textual and visual modalities lead to variations of the resulting images within a range of a single text prompt. For instance, when tasked with rendering a \textit{``low-poly style horse"}, T2I DMs produce various versions of visual outcomes (\fref{fig:text_prompt_not_enough}a). These variations stem from the process of sampling random noise, which complicates the ability to precisely control the outcome. Even with highly detailed text prompts, controlling the exact style of the resulting images is challenging (\fref{fig:text_prompt_not_enough}b). 
In contrast, it would be more beneficial to use images to specify visual elements as visual prompts that cannot be fully conveyed through text description, offering easier control over the style elements in the results (\fref{fig:text_prompt_not_enough}c).

Recent efforts have explored the use of reference images as visual prompts. These approaches include fine-tuning the diffusion model with a set of images containing the same theme \cite{ruiz2023dreambooth, kumari2023multi}, learning new text embeddings \cite{gal2022image, han2023highly, avrahami2023bas}, and adapting cross-attention modules to incorporate  image features \cite{ye2023ip, wang2023styleadapter}. However, these methods necessitate costly fine-tuning and often cause issues such as content from the reference images leaking into the result or inaccurately reflecting the intended style.

In this study, we introduce visual style prompting, a novel approach that guides the desired style using a reference image. As illustrated in \fref{fig:teasor_v0}, our method utilizes a reference image as a visual prompt to extract the specific style and successfully generates images that embody the given style (\eg, \textit{``a pegasus with the nuance of clouds”}). Our key idea lies in maintaining the queries from the original features while swapping the keys and values with those from the reference features. To alleviate the issue of content leakage, we employ this operation only in the latter stages of the self-attention layers. Importantly, our method eliminates the need for fine-tuning and is versatile enough to be implemented in any text-to-image diffusion model.

We first introduce our core contribution, swapping self-attention (\Sref{sec:swapping_selfattention}), and explore its effectiveness in various locations within self-attention blocks in terms of the granularity of visual elements (\Sref{sec:choosing_blocks}). Subsequently, we empirically demonstrate the superiority of our method over existing competitors across various styles and text prompts, evaluating in terms of fidelity to the style, reflection of the text prompt, occurrence of content leakage, and diversity of the generated images (\Sref{sec:exp_comparison_competitors}). Furthermore, we show the successful application of our method to widely used existing techniques (\Sref{sec:exp_existing_tech}) and its effectiveness when using real images as reference images (\Sref{sec:exp_real_image}). Additional results can be found in the Appendix. For research community, we will make our code publicly available.

\section{Visual style prompting}

We propose \ours{} which generates new content conditioned on text prompts with the style of a given reference image without fine-tuning a pre-trained diffusion models. We discuss the attention modules in text-to-image diffusion models and introduce swapping self-attention.

\subsection{Discussion on attention modules in T2I diffusion}
Modern diffusion models consist of a number of self-attention and cross-attention blocks \cite{vaswani2017attention}. Both of them employ the attention mechanism, which first obtains an attention map using similarity between query features $Q$ and key features $K$, then aggregates value features $V$ using the attention map as weights. The attention mechanism is formulated as follows:

\begin{equation}
    \label{eq:attention}
    \text{Attention}(Q, K, V) = \text{Softmax}(\frac{QK^T}{\sqrt{d}})V,
\end{equation}

In T2I DMs, the difference between self-attention and cross-attention is the origin of keys and values. In the cross-attention module, text features extracted by language models serve as the key and value to inject a text prompt as a condition with respect to the query features from the previous layer. On the other hand, in the self-attention module, features with spatial dimensions generated from the previous layer serve as key, query, and value by themselves.

We start our discussion from the simplest way of controlling style: specifying the style in the text prompt. However, ambiguity inherent in text prompts makes it challenging to specify the style precisely as shown in \fref{fig:text_prompt_not_enough}.

Dreambooth variants \cite{ruiz2023dreambooth, kumari2023multi} learn a style by updating the models to be triggered by a unique text identifier. Textual inversion variants \cite{gal2022image, han2023highly} introduce new text embeddings to represent a style. 

Adapter variants \cite{ye2023ip, wang2023styleadapter} introduce extra networks to encode and apply styles of a reference image. All these approaches work with cross-attention which was originally trained for conditioning the denoising process with respect to text prompts. Hence, they are limited to reflecting the styles that were paired with text prompts during training.

\begin{figure}[t]
    \centering
    \includegraphics[width=1.0\linewidth]{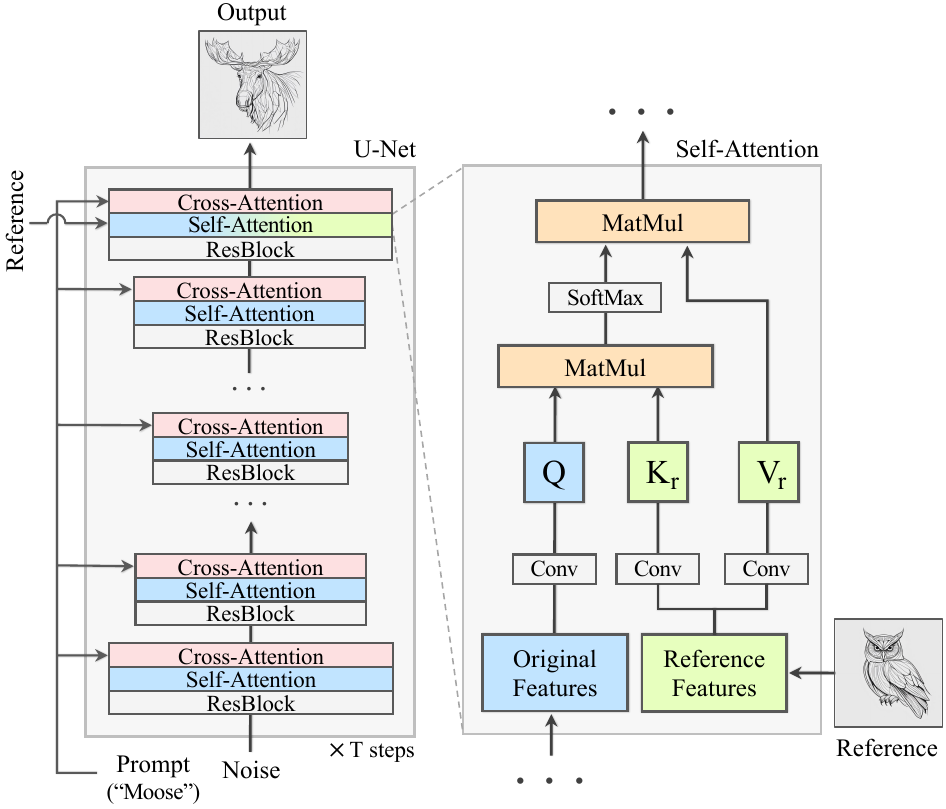}
    \vspace{-3mm}
    \caption{\textbf{Overview of swapping self-attention for visual style prompting.} 
    We swap the key and value features of self-attention block in an original denoising process with the ones from a reference denoising process.
    This procedure is repeated for T steps, 
    resulting in the original content rendered with the style elements from the reference image.
    }
    \label{fig:method_overview}
\end{figure}

On the other hand, self-attention layer receives key and values coming from the main denoising process which has spatial dimensions with more freedom to represent spatially varying visual elements as opposed to the text prompts. As our goal is to reflect style elements from a reference image that are not easily represented by textual description, we opt to borrow key and values of self-attention layers in the reference process to the original process, namely swapping self-attention (\fref{fig:method_overview}).

Interestingly, attention mechanism has been useful in style transfer literature \cite{sheng2018avatar,park2019arbitrary,liu2021adaattn}. They compute similarity between an original image and a reference image in VGG feature maps and reassemble the reference feature maps using the similarity as weights.

\subsection{Swapping self-attention}
\label{sec:swapping_selfattention}

Visual style prompting has two main requirements: reflecting style elements from a reference image and keeping the content specification from a text prompt intact. We assume that we have two denoising processes of style and content, and aim to reflect the style into the original process with swapping self-attention as follows.

We start from two denoising processes: an original process starting from a text prompt and initial noise which will naturally produce an image $I_o$ specified by the text prompt, and a reference process which will end up in a reference image $I_r$. The latter can be prepared from a text prompt specifying a style or an inversion from a real image.

In order to produce an image $I_*$ having the reference style from the original process, we swap the key and value features in the $l^{\text{th}}$ self-attention layer with the key and value features $K_r$ and $V_r$, respectively, from the reference process. Then the attention mechanism computes the similarity between the original query feature $Q$ and the reference key feature $K_r$ to reassemble the reference value feature $V_r$ using the similarity as weights:

\begin{equation}
    \label{eq:attention2}
    \text{Attention}(Q^{l}, K_{r}^{l}, V_{r}^{l}) = \text{Softmax}(\frac{Q^{l}{K_{r}^{l}}^T}{\sqrt{d}})V_{r}.
\end{equation}

\begin{figure}[t]
    \centering
    \includegraphics[width=\linewidth]{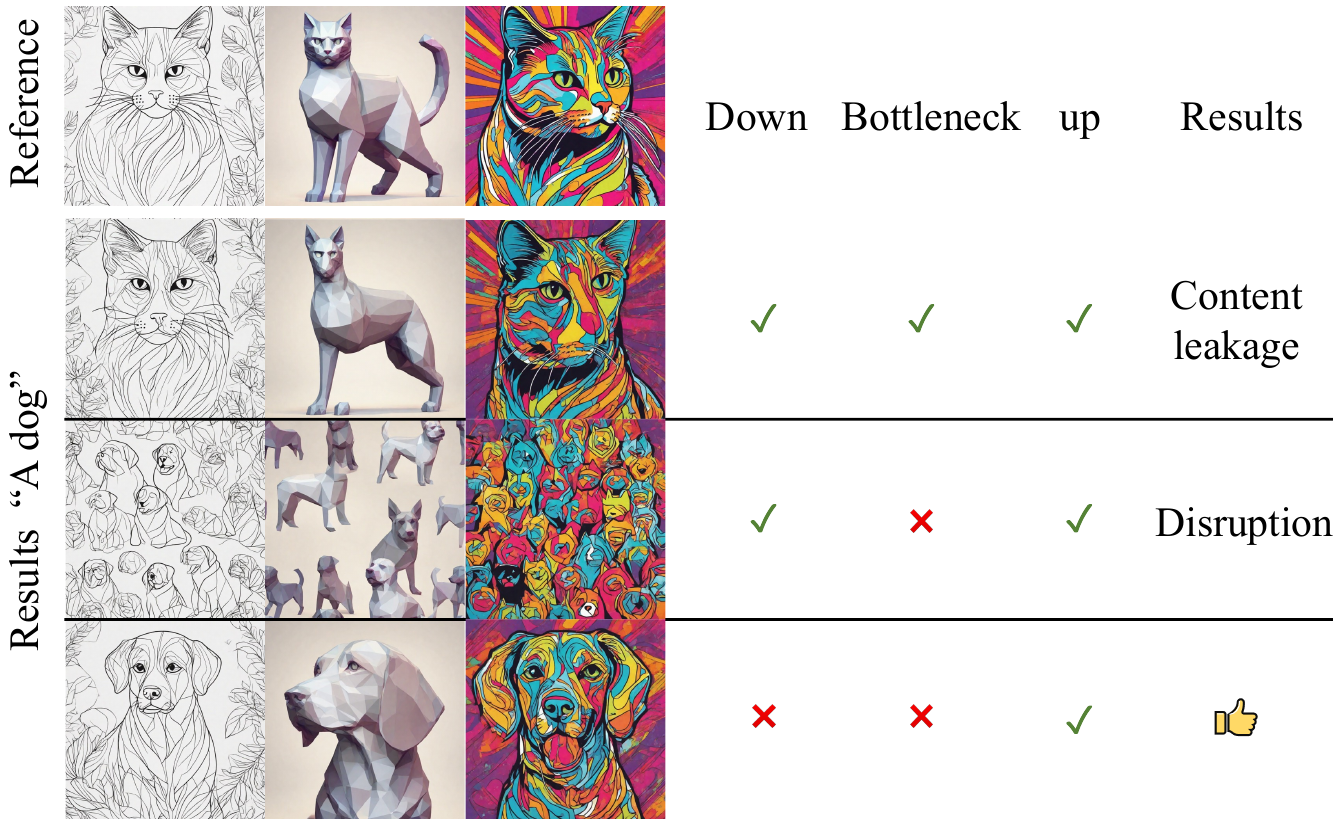}
    \caption{\textbf{The effect of 
    swapping self-attention across different layers.}
    Swapping self-attention on the bottleneck and downblocks causes content leakage producing cat-like results despite of the dog prompt. Swapping self-attention on downblocks produces disrupted results.
    We only apply swapping self-attention in the upblocks to appropriately reflect the style elements.}
    \label{fig:why_only_upblock}
\end{figure}

While swapping self-attention implements the reassembling operation, simply applying to all self-attention layers exposes a content leakage problem, where the content of the reference image influences the resulting image, as shown in the second row of \fref{fig:why_only_upblock}. I.e., the results contain cats even though the prompts specify ``a dog". In the next subsection, we analyze the effect of the position of swapping self-attention.

\begin{figure}[t]
    \vskip 0.2in
    \centering
    \includegraphics[width=0.9\linewidth]{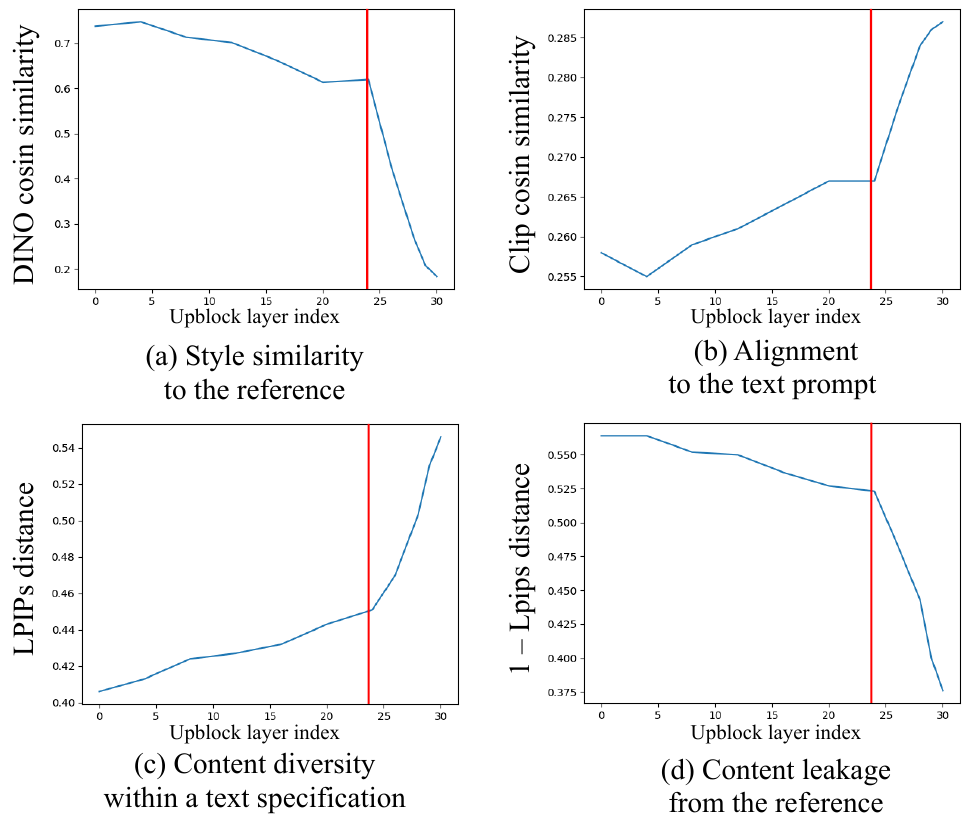}
    \caption{\textbf{Analysis on the optimal range of upblocks for swapping self-attention}.
    We find the optimal range of upblocks for balanced trade-off between different aspects. Please refer to \sref{sec:choosing_blocks} for details.}
    \label{fig:ablation:upblock}
    \vskip -0.2in
\end{figure}

\subsection{Choosing blocks for swapping self-attention}
\label{sec:choosing_blocks}
%\label{sec:selective-attention-blocks}

Here we explore depth of the self-attention blocks to be swapped in the sense of granularity of visual elements.
Modern architecture of diffusion models roughly consists of three sections in a sequence: downblocks, bottleneck blocks, and upblocks. Given that the bottleneck of diffusion models contains content elements of the image~\cite{kwon2023diffusion}, we opt not to apply swapping self-attention to bottleneck blocks to prevent transferring contents in a reference image. The third row of \fref{fig:why_only_upblock} shows that not swapping the bottleneck blocks prevents content leaking from the reference image. However, the synthesized images show disrupted results with seriously scattered objects.

We conjecture that this phenomenon happens because feature maps of downblocks have unclear content layout\cite{cao_2023_masactrl}, so substituting features based on this inaccurate layout causes the disrupted results. To avoid injecting the unneccesary features, we choose to swap the key and value of self-attention only in upblocks. The last row of \fref{fig:why_only_upblock} shows the success of our strategy.

Since recent large T2I DMs consist of many blocks, we further analyze the behavior by changing the start of the swapping with the end of the swapping fixed at the end.
We use four key metrics: (a) style similarity to the reference image, (b) alignment to the text prompt, (c) content diversity within a text specification, and (d) content leakage from the reference image. As shown in \fref{fig:ablation:upblock}, there is a point where all four metrics abruptly change (red line). We choose this point as the optimal starting point with balanced trade-off among all aspects. We provide qualitative results with detailed split of layers in \fref{afig:layer_ablation_qual_anime} and \dref{afig:layer_ablation_qual_lie_art}

\begin{figure}[t]
    \centering
    \includegraphics[width=0.8\linewidth]{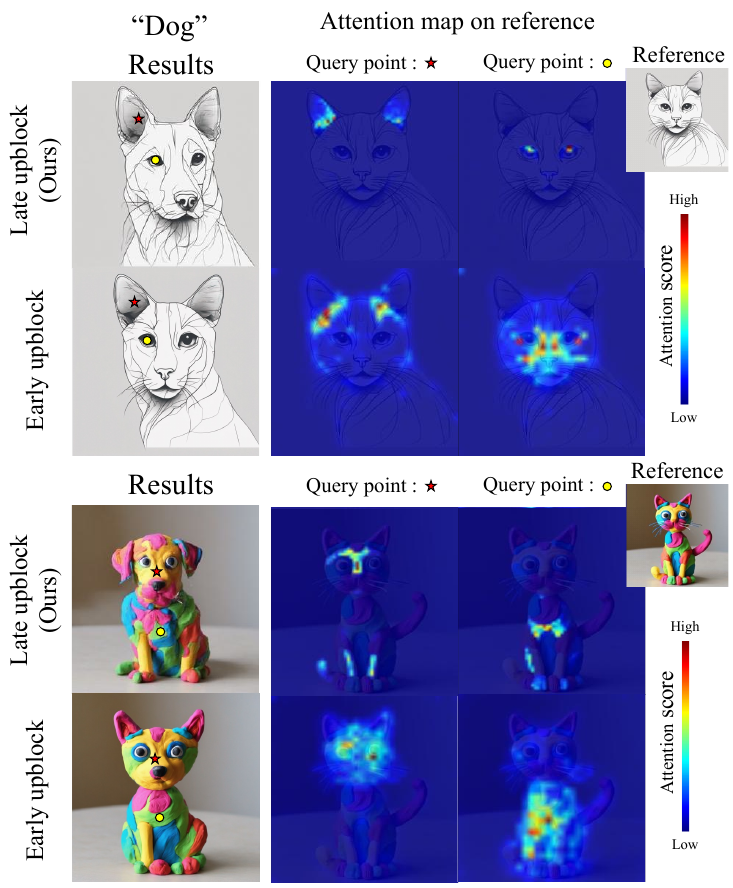}
    \caption{\textbf{Attention map visualization over late and early upblock layers.}
    The late upblock better focuses on the semantically corresponding region than the early upblock, leading to more freedom to reassemble small parts. The early upblock attends larger region leading to content leakage.
    }
    \label{fig:ablation:attentionmap}
\end{figure}

\fref{fig:ablation:attentionmap} compares average attention maps from the late upblock and the early upblock applying swapping self-attention. 
Using late upblock has more freedom to reassemble the reference style elements leading to more doggy results than early upblock which produces some cat-like attributes.
The right two columns visualize the attention weight of query points marked as red stars and yellow dots. 
% Swapping self-attention on late upblock reassembles features from a semantic correspondence, e.g., ear and nose. 
Swapping self-attention on late upblock reassembles features from a style correspondence, e.g., texture and color.
On the other hand, swapping self-attention on early upblock reassembles features from wider area with different styles. This comparison clarifies the reasons for using only late upblock. Please refer to \fref{afig:layer_ablation_visual_advanced} for a detailed analysis.

\begin{figure*}[t]
    \centering
    \includegraphics[width=\textwidth]{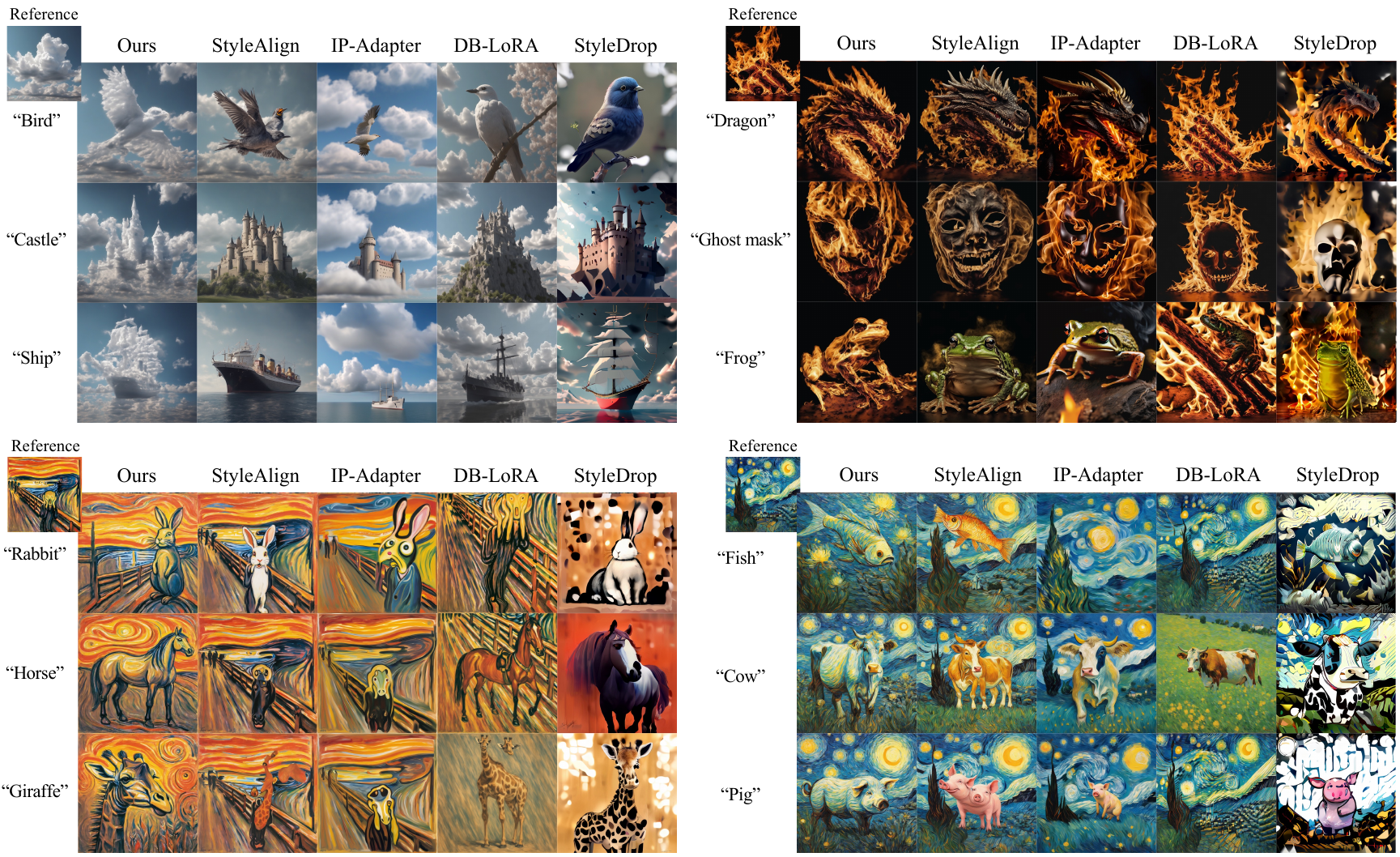}
    \caption{\textbf{Qualitative comparison across various styles and text prompts.} \ours{} successfully reflects style elements in reference images while not causing content leakage from the reference images.}
    \label{fig:vs_competitor}
\end{figure*}

\section{Experiments}
\paragraph{Details} 
We use SDXL \cite{podell2023sdxl} as our pretrained text-to-image diffusion model and choose the $\text{24}^\text{th}$ layer and the after. We also validate our methods on Stable diffusion (SD) v1.5 \cite{rombach2022high}. Results with SD v1.5 is shown in \fref{afig:SDv15}. We set classifier-free guidance as 7.0 and run DDIM sampling with 50 timesteps following the typical setting. 
The initial noises for a text prompt are identical across competitors for fair comparison.
We use official implementation of IP-Adapter and StyleAlign. For Dreambooth-LoRA\footnote{https://huggingface.co/docs/diffusers/training/lora}
, we use diffusers pipeline provided by Huggingface. Since official code of StyleDrop is not available, we use an unofficial implementation\footnote{https://github.com/huggingface/diffusers/tree/main/examples/amused} provided by Huggingface. All competitors are based on SDXL except StyleDrop (unofficial MUSE).
As Dreambooth-LoRA, a training-based approach, requires multiple images, we train the models with five images: the original image and quarter-split patches of the reference image, because using only one image usually leads to destructive results or suffers from overfitting.
For IP-adapter, we choose $\lambda{}=0.5$ which is the best weighting factor for the task. For the visualization of the attention maps in \fref{fig:ablation:attentionmap}, we average the multi-head attention maps all together along the channel axis at 20th denoising timestep.

\paragraph{Metrics}
Following \cite{voynov2023p+, ruiz2023dreambooth}, we use DINO (ViT-B/8) embeddings \cite{caron2021emerging} to measure style similarity between a reference image and a resulting image. We use CLIP (ViT-L/14) embeddings \cite{radford2021learning} to measure the alignment between text prompts and resulting images. We use LPIPS metric \cite{zhang2018unreasonable} to measure diversity by average LPIPS between different resulting images in the same text prompt.
For quantitative evaluation and comparison, we prepare 720 synthesized images from 40 reference images, 120 content text prompts (3 contents per 1 reference), and 6 initial noises. The reference images are generated from 40 stylish text prompts. \aref{asec:style-content-prompt-list} provides the text prompt set.

 \begin{figure}[t]
    \centering
    \includegraphics[width=0.8\linewidth]{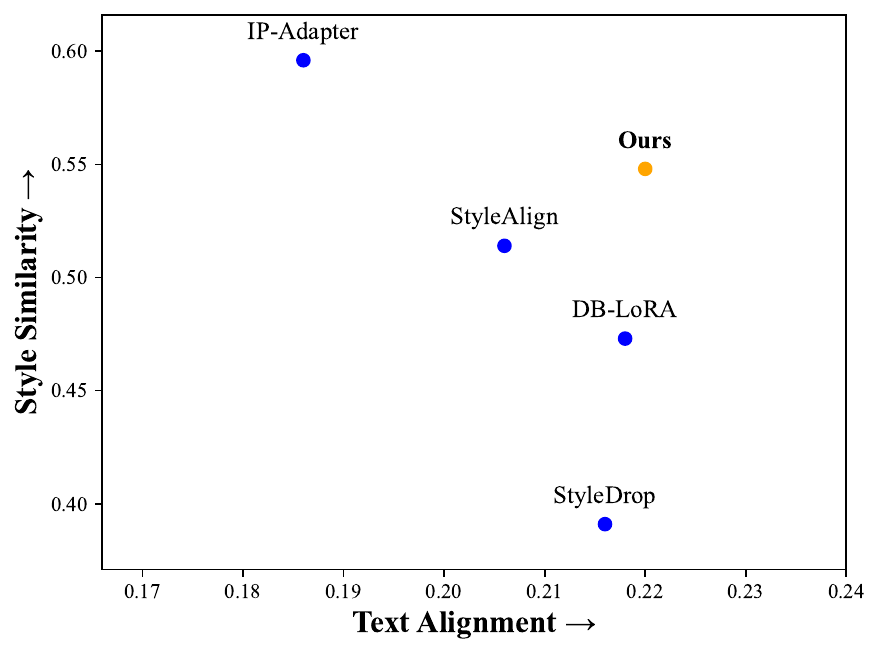}
    \caption{\textbf{Quantitative comparison}. We compare the results for text alignment (CLIP score) and style similarity (DINO embedding similarity) between other methods (blue points) and our method (orange point).}
    \label{fig:vs_competitors_quan}
\end{figure}

\subsection{Comparison against competitors}
\label{sec:exp_comparison_competitors}

We compare our method with StyleAlign \cite{hertz2023style}, IP-Adapter \cite{ye2023ip}, Dreambooth-LoRA \cite{ruiz2023dreambooth,lora_repo}, and StyleDrop \cite{sohn2023styledrop}.

\paragraph{Style \& content control}
We provide a qualitative comparison with competitors in \fref{fig:vs_competitor}. We focus on controlling the two main aspects: style \& content. Our method synthesizes the content specified by the text prompt using the style only from the reference image. On the other hand, other methods infuse color or texture that are not present in the reference such as genuine feathers, bricks, iron, or skin. In addition, they often struggle with content leakage from the reference such as the same layout, screaming person, or a castle. It harms faithfulness to the text prompt. Quantitative results in \fref{fig:vs_competitors_quan} agrees on above statements: 
IP-Adapter shows higher style similarity but it seriously neglects text prompts.

\begin{table}[t]
\vskip 0.1in
\centering
\caption{
\textbf{User study comparison.} We asked participants: Which method best reflects the style in the reference image AND the content in the text prompt?
}
\small
\begin{tabular}{ccccc}
\toprule
Ours & StyleAlign & IP-Adapter & DB-LoRA & StyleDrop \\
\midrule
\textbf{58.15}\% & 13.15\% & 18.47\% & 7.66\% & 2.58\% \\
\bottomrule
\end{tabular}
\label{tab:userstudy}
\end{table}

\paragraph{User study} 
For more rigorous evaluation, we conducted a user study with 62 participants. We configured a set with a reference image, a content text prompt, and six synthesized images with different initial noises per method from five competitors. The participants answered below question for 20 sets: Which method best reflects the style in the reference image AND the content in the text prompt?
As indicated in \tref{tab:userstudy}, the majority of participants rated our method as the best.
The lower ratings of the IP-Adapter in the user study may be attributed to its poor text alignment, as illustrated in \fref{fig:vs_competitors_quan} despite its high style similarity.
Examples of the user study are provided in \fref{afig:user_study_ex}.

\begin{figure}[t]
    \centering
    \includegraphics[width=\linewidth]{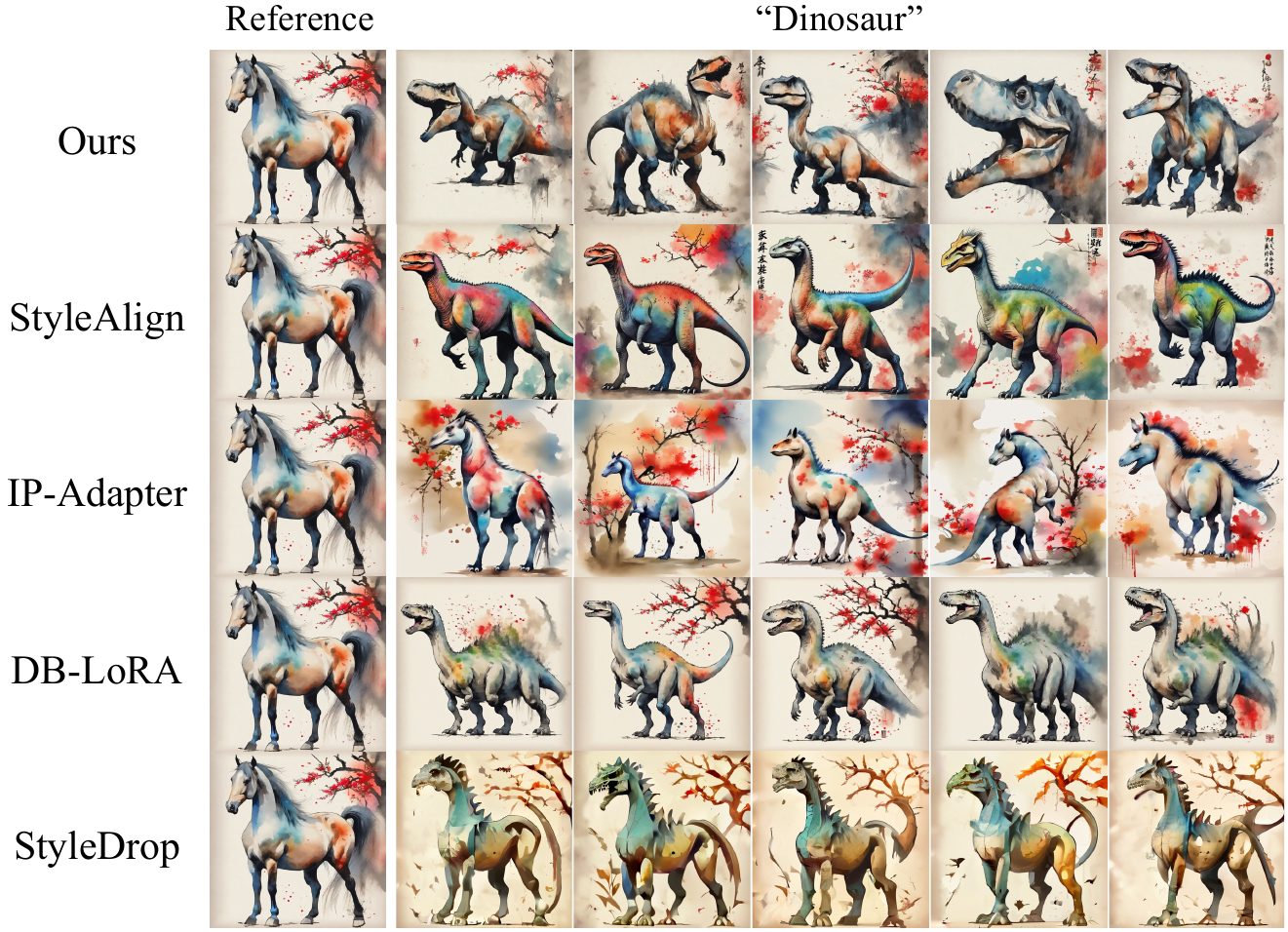}
    \caption{\textbf{Qualitative comparison across same style.} Competitors face challenges in generating images with diverse layouts and compositions due to content leakage from the reference.}
    \label{fig:diversity_vs_competitors}
\end{figure}

\begin{figure}[t]
    \centering
    \includegraphics[width=\linewidth]{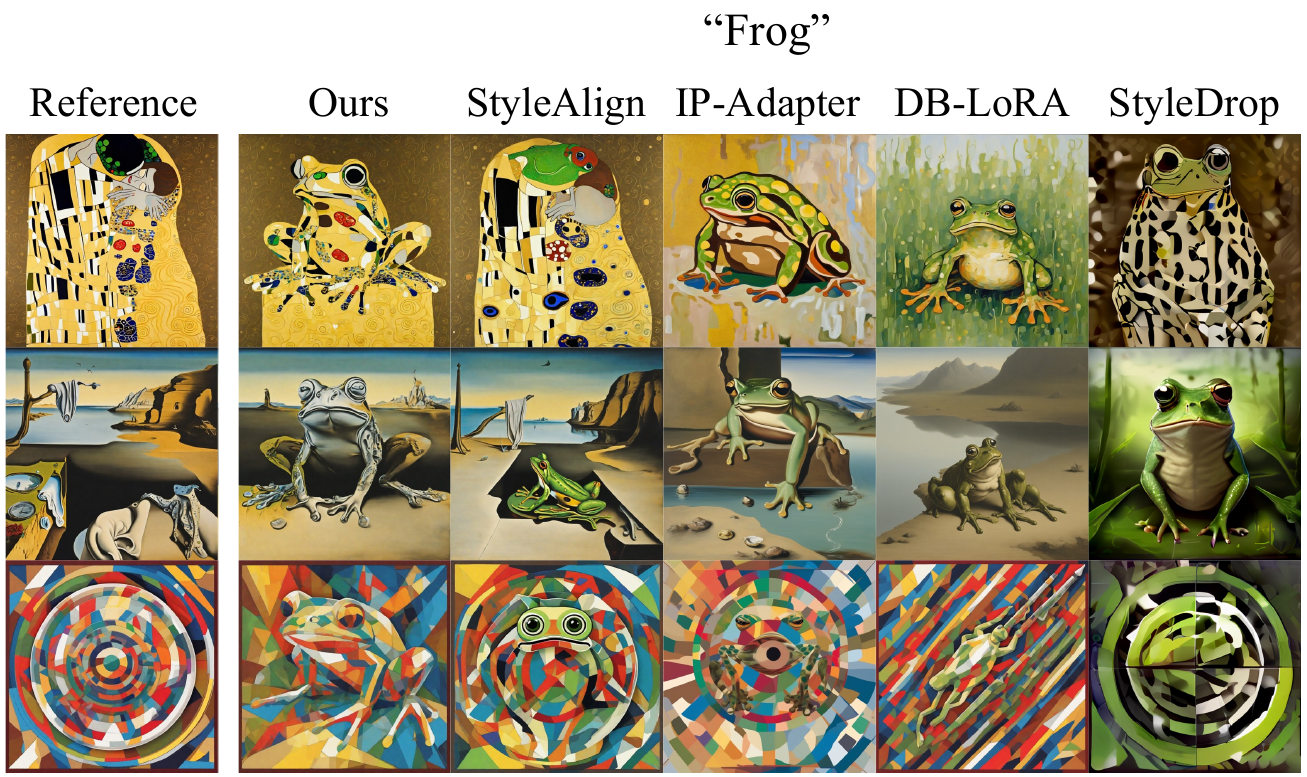}
    \caption{\textbf{Qualitative comparison with varying styles and fixed content.} \Ours{} reflects style elements from various reference images to render ``frog" while others struggle.}
    \label{fig:vs_competitor_frog}
\end{figure}

\paragraph{Diversity within a text specification}
Starting from different initial noises, the diffusion models trained on a large dataset produce diverse results within a specification of a text prompt. \fref{fig:diversity_vs_competitors} shows that our results have various poses and viewpoints while others barely change, i.e., other methods limit the diversity of the pre-trained model. \fref{afig:more_diversity_results} provides more examples. 

\begin{figure}[t]
    \centering
    \includegraphics[width=0.9\linewidth]{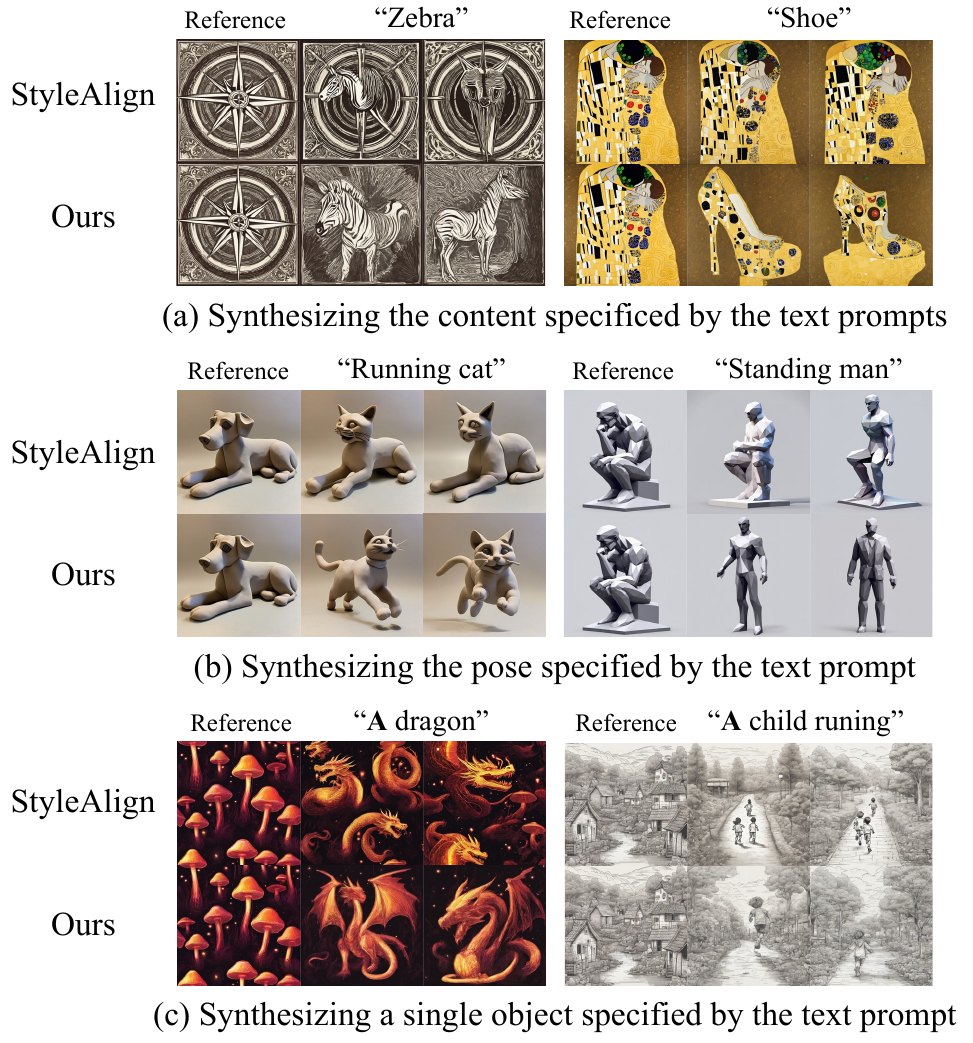}
    \caption{
    \textbf{Comparison for content leakage.} While StyleAlign suffers from content leakage from the reference, results from ours clearly align with the text prompts
    }
    \label{fig:ablation:vs_SA_senario}
\end{figure}

\begin{figure}[t]
    \centering
    \includegraphics[width=0.9\linewidth]{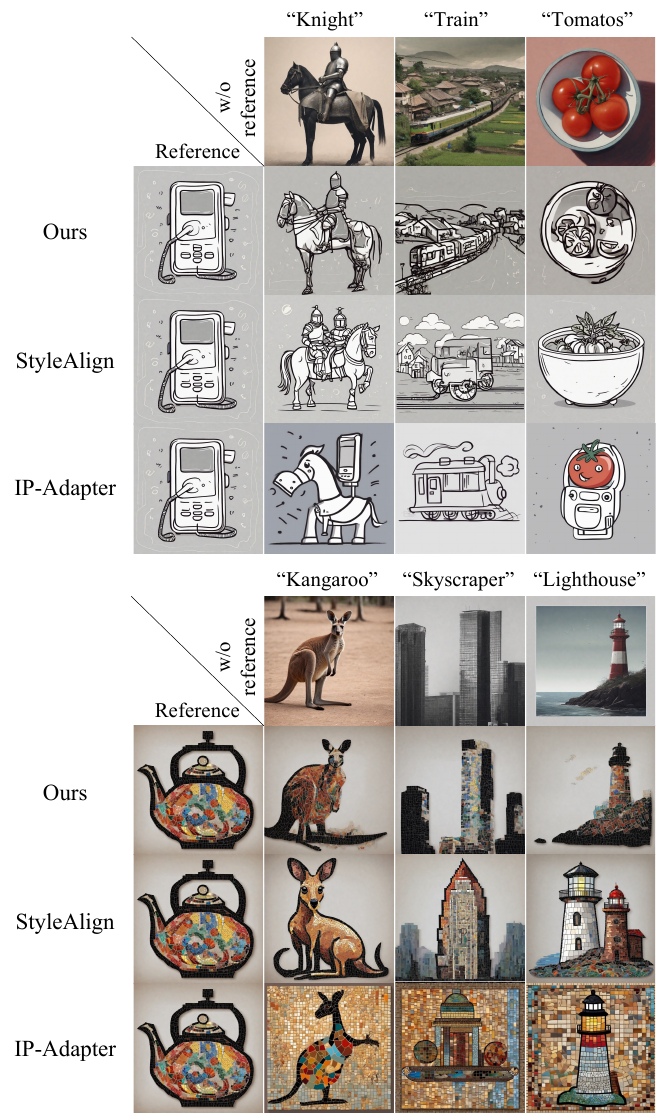}
    \caption{\textbf{Comparison of contents change while reflecting the style in the reference.} Each column shares the same initial noise. \Ours{} reflects the style in the reference with minimal changes in contents of the original denoising process. } 

    \label{fig:ablation:init_noise_is_content}
\end{figure}

\paragraph{Content leakage}
Content leakage refers to the phenomenon where the content of a reference image appears in a result. As evident in \fref{fig:vs_competitor_frog}, our model exhibits significantly less content leakage when compared to other models. We focus on a comparison against a runner-up method, StyleAlign, particularly in terms of how content leakage can be an obstacle to controlling the content using a text prompt.

\fref{fig:ablation:vs_SA_senario}a compares examples where strong content leakage prevents the content from the text from appearing in StyleAlign. We often observe famous paintings in the reference easily leak into the results of StyleAlign while ours does not struggle. \fref{fig:ablation:vs_SA_senario}b compares examples where the pose in the reference leaks into the results of StyleAlign preventing the reflection of specified pose in the text prompt. On the other hand, our method reflects the correct poses. \fref{fig:ablation:vs_SA_senario}c compares examples where the number of small instances in the reference leaks into the result of StyleAlign. Contrarily, our method correctly synthesizes a single instance of the content specified by the text prompt.

\paragraph{Preserving the content of the original denoising process}
\fref{fig:ablation:init_noise_is_content} shows the results of ours and other methods using the same initial noise in each column. Our method precisely reflect the style in the reference with minimal changes of contents in the original denoising process. On the other hand, the other methods severely alters pose, shape, or layouts. It is an important virtue of controlling style to keep the rest intact.

\subsection{Combining existing techniques}
\label{sec:exp_existing_tech}
\begin{figure}[t]
    \vskip 0.1in
    \centering
    \includegraphics[width=1.0\linewidth]{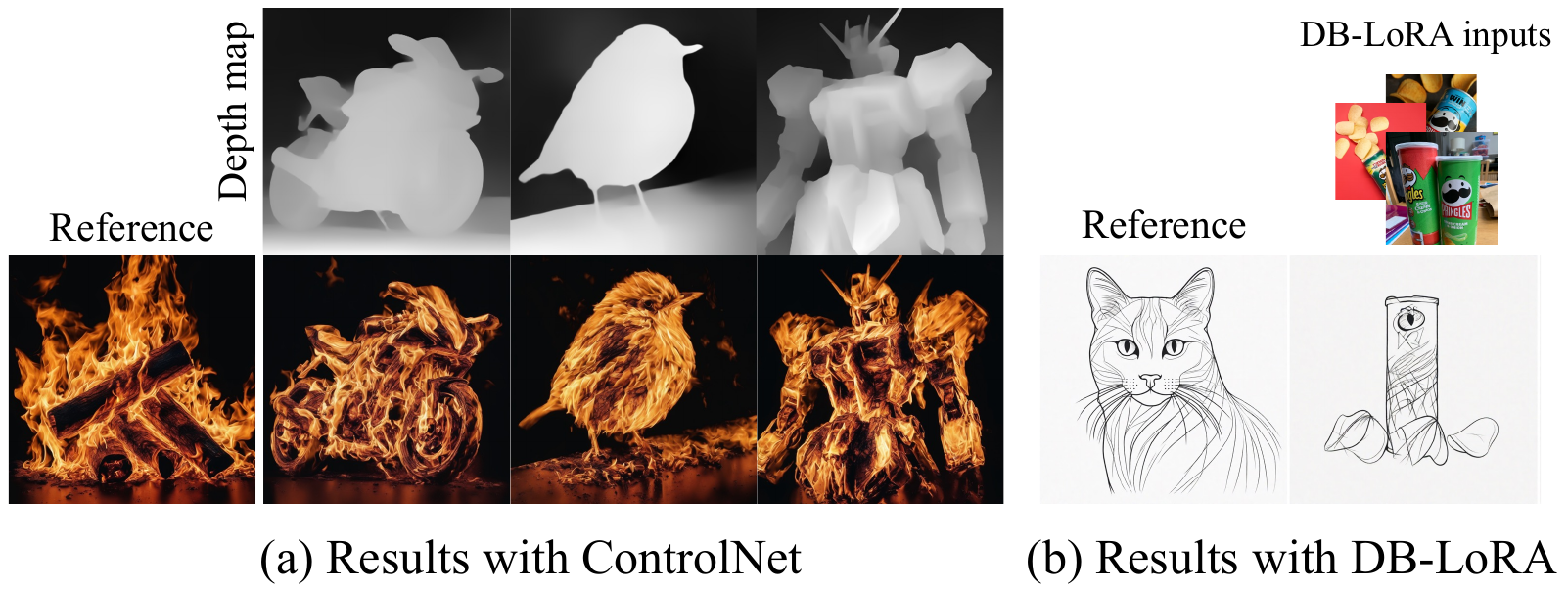}
    \caption{\textbf{Visual style prompting with existing techniques.} Our method is compatible with ControlNet and Dreambooth-LoRA.}
    \label{fig:controlnet_db_lora}
\end{figure}

\Ours{} is compatible with famous techniques such as ControlNet \cite{zhang2023adding} and Dreambooth-LoRA \cite{lora_repo} as shown in \fref{fig:controlnet_db_lora}. It allows more specific control of the results such as following depth maps or inserting custom contents.

\subsection{Real image as a reference image}
\label{sec:exp_real_image}
\begin{figure}[t]
    \centering
    \includegraphics[width=0.9\linewidth]{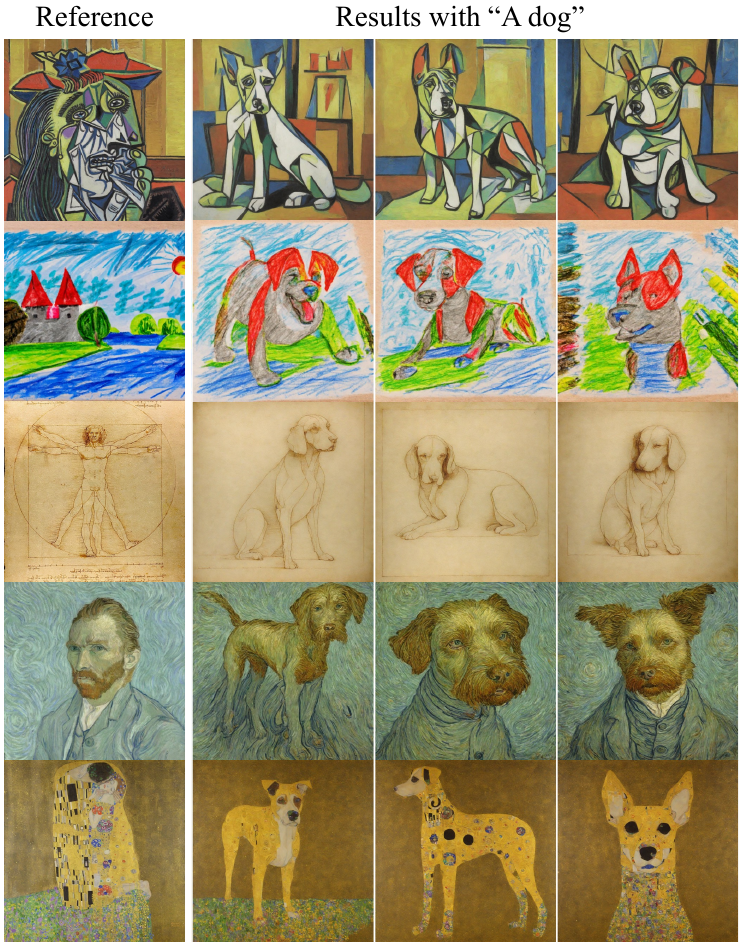}
    \caption{
    \textbf{Visual style prompting with real images as references.}
    \Ours can reflect the style of real images (left) with inversion, still representing the content of text prompts.}
    \label{fig:ablation:real_img}
\end{figure}

\fref{fig:ablation:real_img} shows that \ours{} can take real images as style reference by inverting them via diffusion process. For the reference denoising process, we use the name of the famous paintings or simple descriptions as text condition.
While we empirically choose stochastic inversion rather than DDIM inversion \cite{song2020denoising}, designing better inversion methods would be a good future research.

\section{Related work}

\subsection{Customization of T2I diffusion models}
Diffusion models~\cite{ho2020denoising, song2021scorebased} open new era in image synthesis. Especially, models trained with extensive text and image paired data~\cite{saharia2022photorealistic, rombach2022high, ramesh2022hierarchical} show impressive text-to-image conversion performance. There has been a growing interest in generating custom objects or styles. Dreambooth~\cite{ruiz2023dreambooth, kumari2023multi, everaert2023diffusion} and LoRA~\cite{lora_repo, zhang2023adding, li2023gligen} variants fine-tune a pre-trained diffusion model with a few images containing the desired object or style. Textual inversion variants~\cite{gal2022image, han2023highly} learn a customized text embedding using iterative optimization. Additionally, Adapter variants \cite{ye2023ip, wang2023styleadapter} aim to convert the visual feature of CLIP into cross-attention embedding of DMs by exploiting general large image datasets with different styles. However, all these methods require additional training. The training process for each style or model impedes practical usage.

\subsection{Content leakage in style customization}
Methods for customization~\cite{ruiz2023dreambooth, lora_repo, sohn2023styledrop} can be used to apply custom style to diffusion models by training a style identifier text composed with a text prompt describing contents (e.g. A cat in style of S*). However, they often suffer content leakage problem, which reflects undesired content of a reference image on resulting images. To address the problem, StyleDrop \cite{sohn2023styledrop} employs 2-stage fine-tuning with iterative feedback, and StyleAdaptor \cite{wang2023styleadapter} shuffles positional embeddings of CLIP encoder, removes class token, and uses multiple reference images. Still, describing images in a specific format such as contents followed by style phrase is inconvenient. 

StyleAlign \cite{hertz2023style} introduces shared self-attention which merges keys and values of self-attention layers from an original and a reference processes. Existence of the keys from the original process hinders the reference from squeezing into the original process leading to less reflection of the reference style. \fref{afig:vs_SA_attn} shows that shared self-attention takes both keys into account. Furthermore, it requires AdaIN \cite{huang2017arbitrary} operation to normalize queries and keys of the original process using the queries and keys of the reference process. We suspect that it changes the content of resulting images different from the original process, leading to less alignment to the text prompt. On the other hand, our method does not require such preprocessing before swapping self-attention. \aref{afig:vs_SA_definition_of_style} provides more comparison.

\subsection{Controlling diffusion models with visual prompt}
Research exploring the control of diffusion models using visual prompts is prevalent. Among them, image-to-image translation task provides conditional information in the form of an image, edges, semantic maps, and depth maps. ControlNet~\cite{zhang2023adding} stands out as one of the most effective frameworks to guide structural contents. In addition, multiple works~\cite{vsubrtova2023analogies,sun2023imagebrush,nguyen2023visual} propose frameworks that solve image analogy where instructions demonstrating the desired manipulation. However, a pair of the original and the edited image is required.

\subsection{Using self-attention features in diffusion models}
Plug-n-Play\cite{tumanyan2023plug} injects self-attention features to maintain the structure of the given image. In \cite{cao_2023_masactrl}, self-attention is utilized to the preservation of identity during semantic editing within the same instance. In \cite{wu2023tune, yang2023rerender}, sharing self-attention features improves temporal consistency over multiple frames.

\section{Conclusion and Limitation}

In this paper, we introduce \ours{} using \swappingselfattention, which effectively prompts the style of the reference images without content leakage in a training-free manner. We show that using a self-attention block is a better strategy compared to the other methods using a cross-attention block. In addition, we provide a principled way of choosing an optimal range of \swappingselfattention{} with quantitative measurements. Finally, \ours{} qualitatively \& quantitatively outperforms the existing methods. Since \ours{} is limited to the ability of the pre-trained diffusion models, it is impossible to synthesize what the models can not generate and the quality of the resulting images depends on the performance of the models. In addition, due to our method's strong adherence to the style of reference images, it tends to ignore the style specified in the text when both are provided. 

For future work, (1) exploring inversion methods of a real image that are highly compatible with \ours{}; and (2) extending our method to other domains, such as applying it to video content will be a good research direction for broadening its applicability and utility.

\vspace{5pt}

\section{Impact statements}
 Generative models in computer vision are widely used for generating novel visual content. However, ethical concerns regarding copyright infringement and misrepresentation of original artworks need to be addressed. The societal impact, including its influence on cultural perceptions and artistic expression, requires careful consideration. Establishing ethical guidelines and regulations is crucial for its responsible use.

\paragraph{Acknowledgement}
All experiments were conducted on NAVER Smart Machine Learning (NSML) platform \cite{sung2017nsml,kim2018nsml}.

\nocite{langley00}

\clearpage
% \bibliography{example_paper}
\bibliography{main}
\bibliographystyle{icml2024}

% APPENDIX

%%%%%%%%%%%%%%%%%%%%%%%%%%%%%%%%%%%%%%%%%%%%%%%%%%%%%%%%%%%%%%%%%%%%%%%%%%%%%%%
%%%%%%%%%%%%%%%%%%%%%%%%%%%%%%%%%%%%%%%%%%%%%%%%%%%%%%%%%%%%%%%%%%%%%%%%%%%%%%%
% APPENDIX
%%%%%%%%%%%%%%%%%%%%%%%%%%%%%%%%%%%%%%%%%%%%%%%%%%%%%%%%%%%%%%%%%%%%%%%%%%%%%%%
%%%%%%%%%%%%%%%%%%%%%%%%%%%%%%%%%%%%%%%%%%%%%%%%%%%%%%%%%%%%%%%%%%%%%%%%%%%%%%%
\newpage
\appendix
\onecolumn
\section{Appendix.}

\renewcommand{\thetable}{A\arabic{table}}
\renewcommand{\thefigure}{A\arabic{figure}}
\setcounter{figure}{0}
\setcounter{table}{0}

\begin{figure*}[h]
    \centering
    \includegraphics[width=0.8\textwidth]{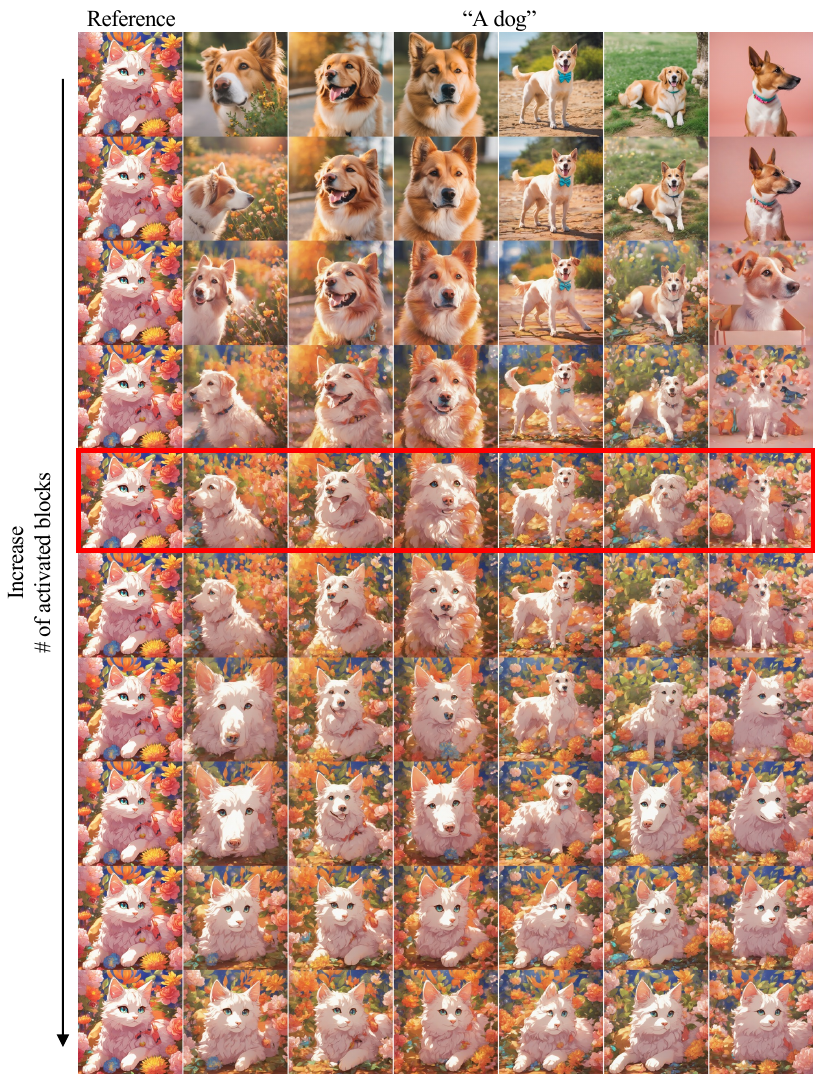}
    \caption{Selective \crossstyleattention{} is important to avoid content leakage while preserving style similarity. Content leakage decreases diversity and text alignment.}
    \label{afig:layer_ablation_qual_anime}
\end{figure*}

\begin{figure*}
    \centering
    \includegraphics[width=0.8\textwidth]{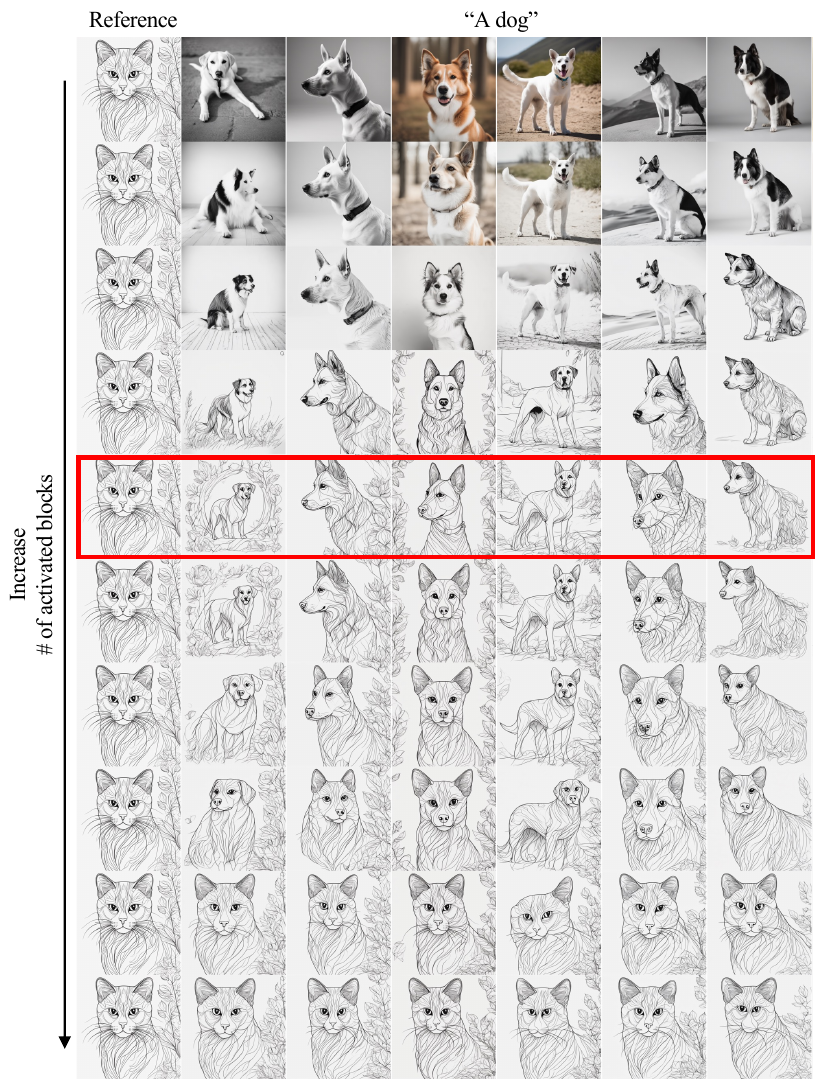}
    \caption{Selective \crossstyleattention{} is important to avoid content leakage while preserving style similarity. Content leakage decrease diversity and text alignment.}
    \label{afig:layer_ablation_qual_lie_art}
\end{figure*}

\begin{figure}[h]
    \centering
    \includegraphics[width=0.8\textwidth]{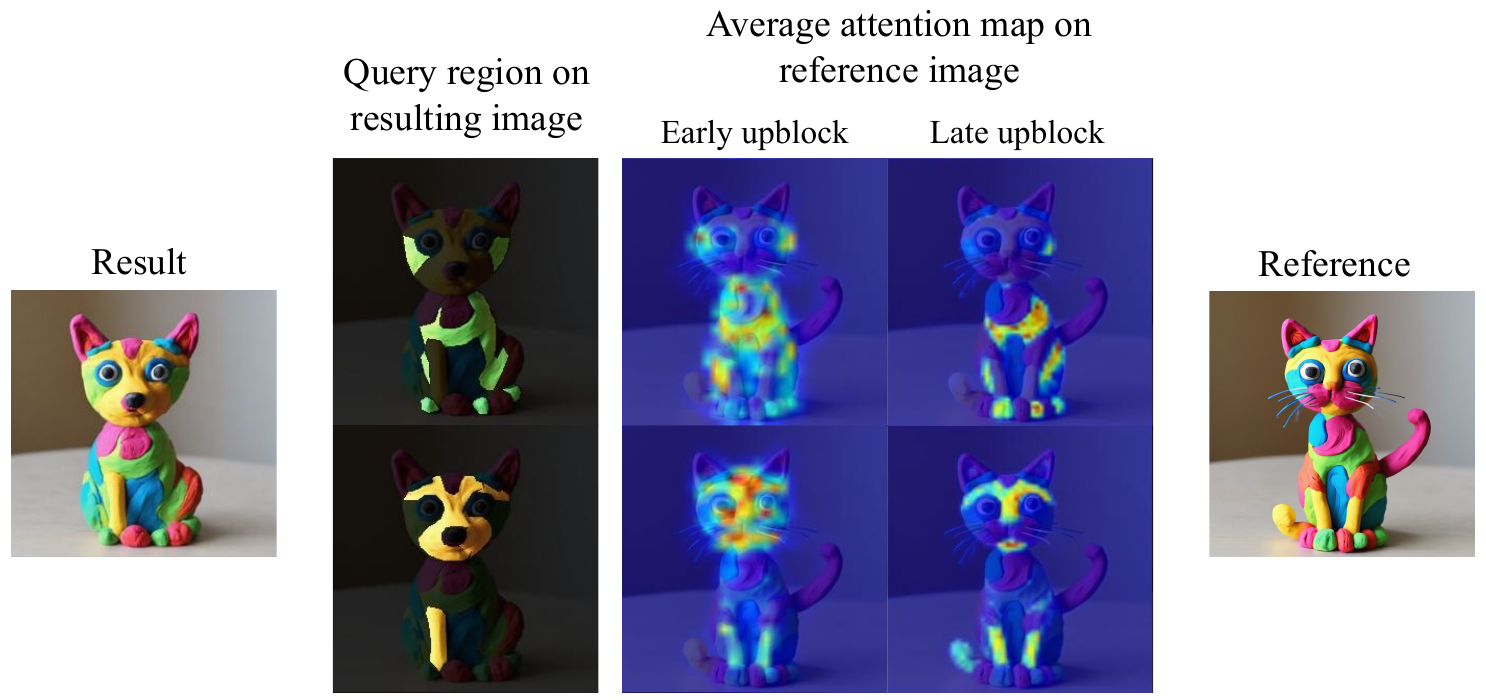}
    \caption{In \fref{fig:ablation:attentionmap}, we only show attention maps of 2 query points. Here, we provide the average attention map of multiple query points on the corresponding query region. At the late upblock, the query point region of the resulting image corresponds to the same style region of the reference image. On the other hand, at the early upblock, the query point region matches not only with the corresponding style region but also with the wider region}
    \label{afig:layer_ablation_visual_advanced}
\end{figure}

\begin{figure}[h]
    \centering
    \includegraphics[width=0.8\textwidth]{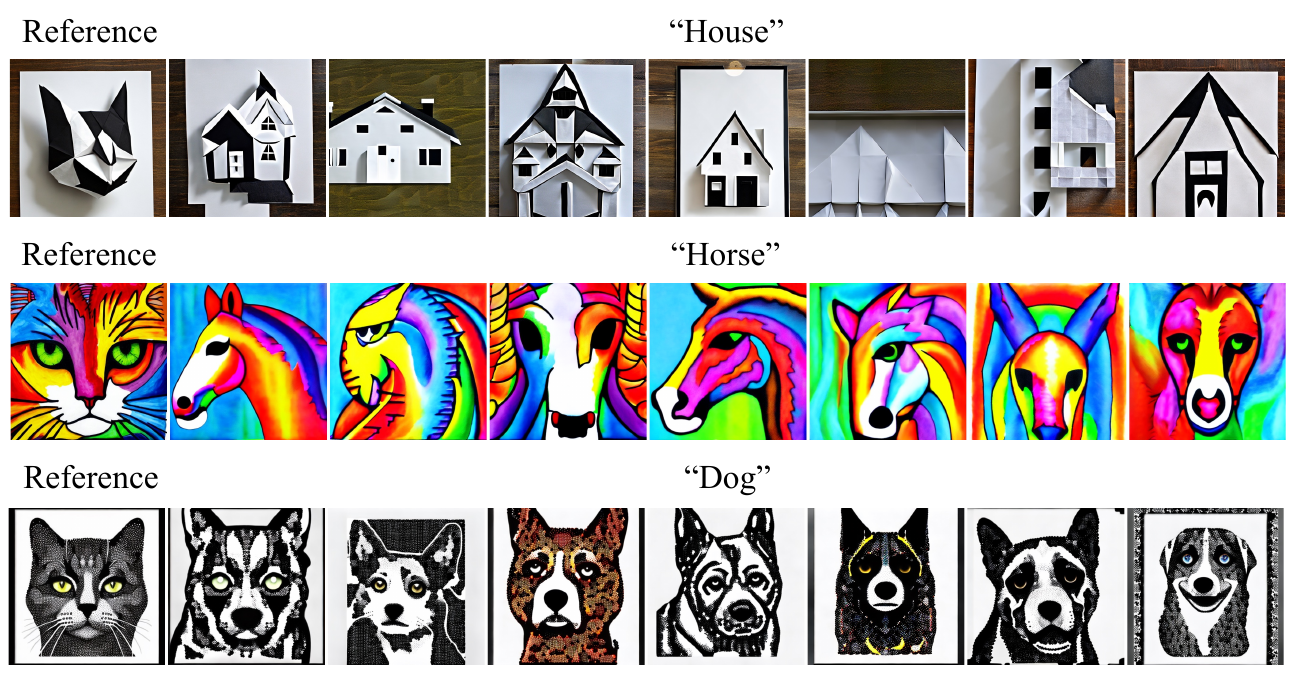}
    \caption{\textbf{Qualitative result of \ours{} on stable diffusion v1.5} Ours also works on the other pretrained diffusion models.}
    \label{afig:SDv15}
\end{figure}

\begin{figure*}
    \centering
    \includegraphics[width=1.0\textwidth]{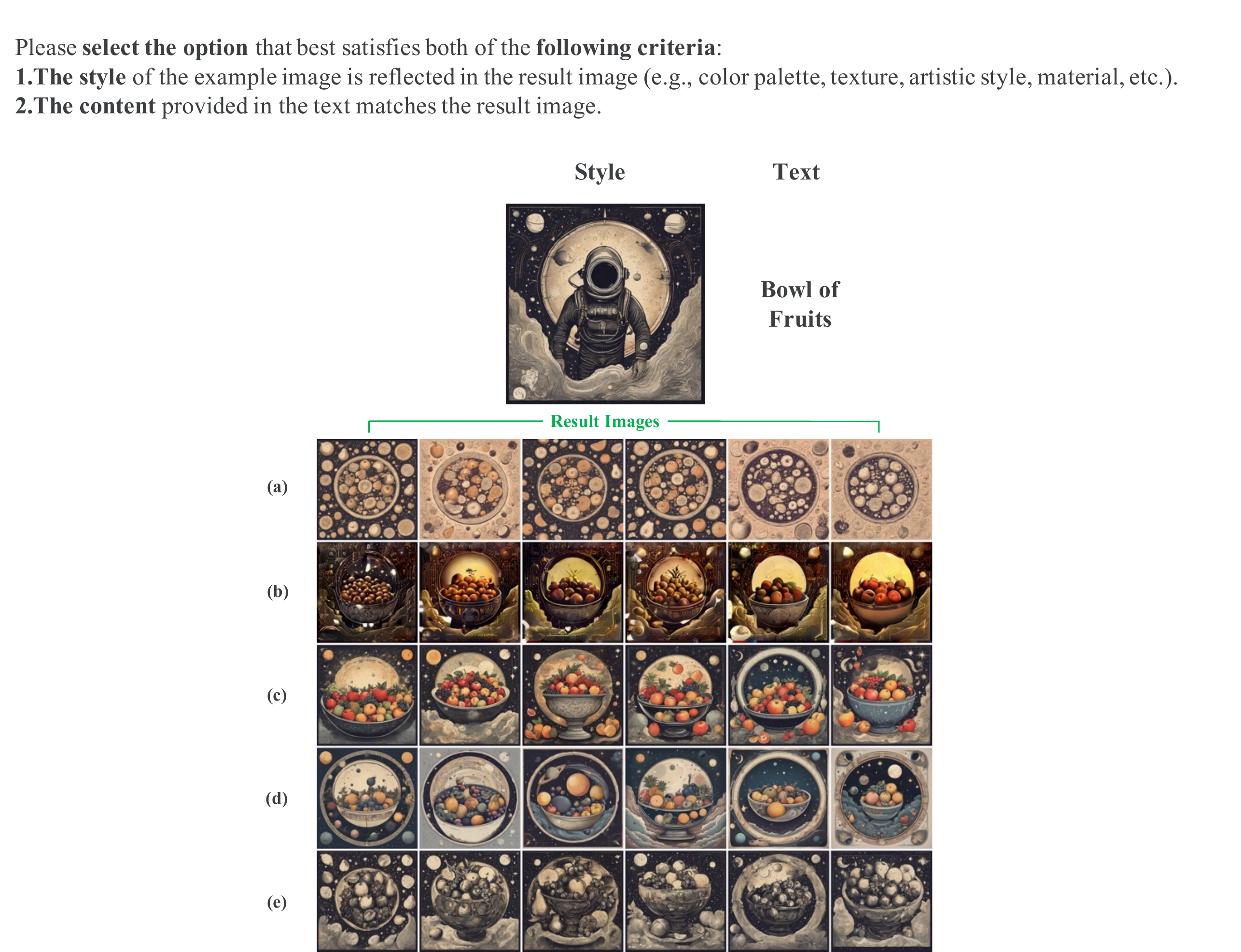}
    \caption{\textbf{Example of a user study.} Each row of images represents the result obtained by different method. The user had to assess
which row is better in terms of style alignment and text alignment.}
    \label{afig:user_study_ex}
\end{figure*}

\begin{figure}[t]
    \centering
    \includegraphics[width=0.9\textwidth]{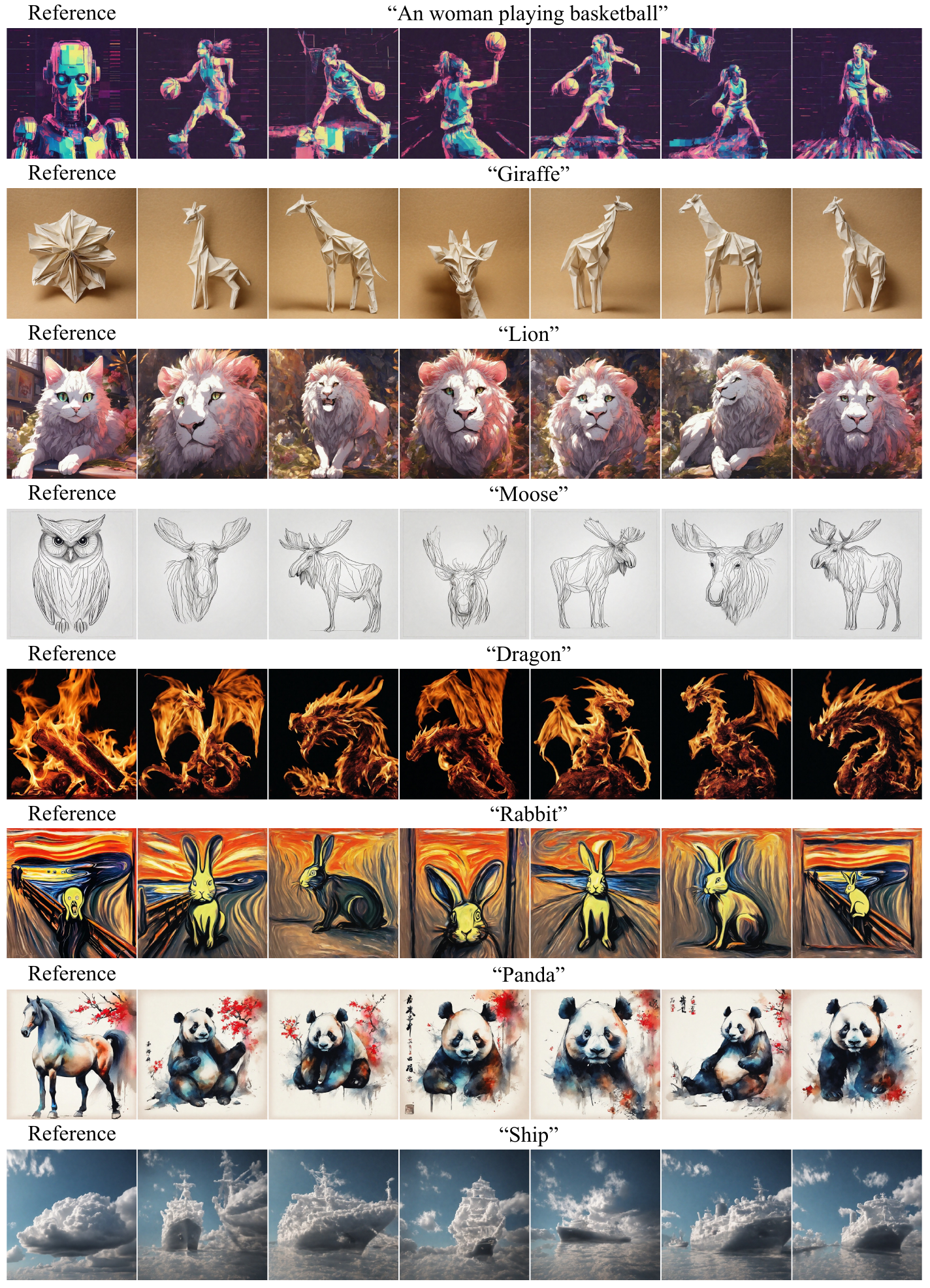}
    \caption{\textbf{Qualitative result of \ours{} within a prompt.} Ours can generate diverse layouts, poses and composition within a prompt.}
    \label{afig:more_diversity_results}
\end{figure}

\begin{figure}[h]
    \centering
    \begin{minipage}[t]{0.48\textwidth}
        \centering
        \includegraphics[width=\textwidth]{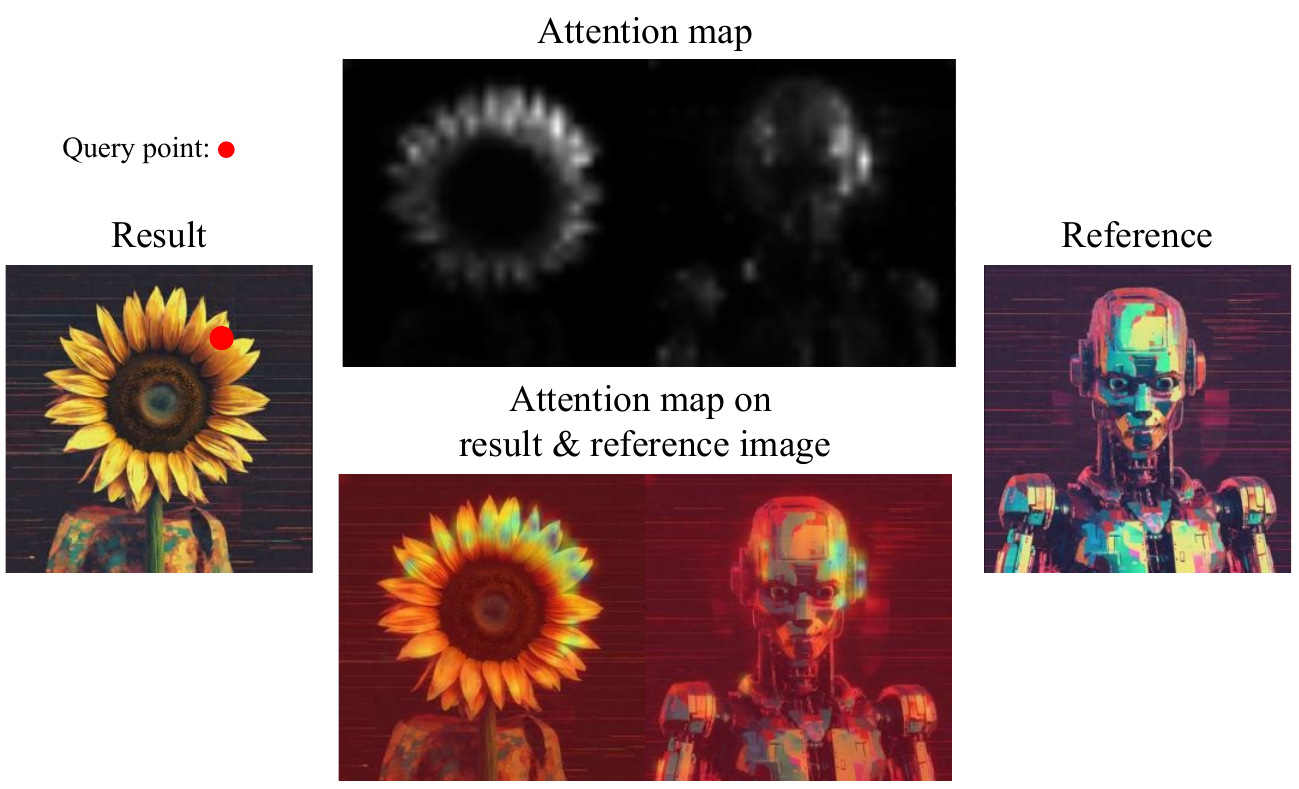}
        \caption{StyleAlign attends both on a reference and a resulting image for shared self-attention mechanism. In contrast, \ours{} only attends on a reference features which leads to better reflection of style in the reference image.}
        \label{afig:vs_SA_attn}
    \end{minipage}
    \hfill
    \begin{minipage}[t]{0.48\textwidth}
        \vspace{-250pt}
        \includegraphics[width=\textwidth]{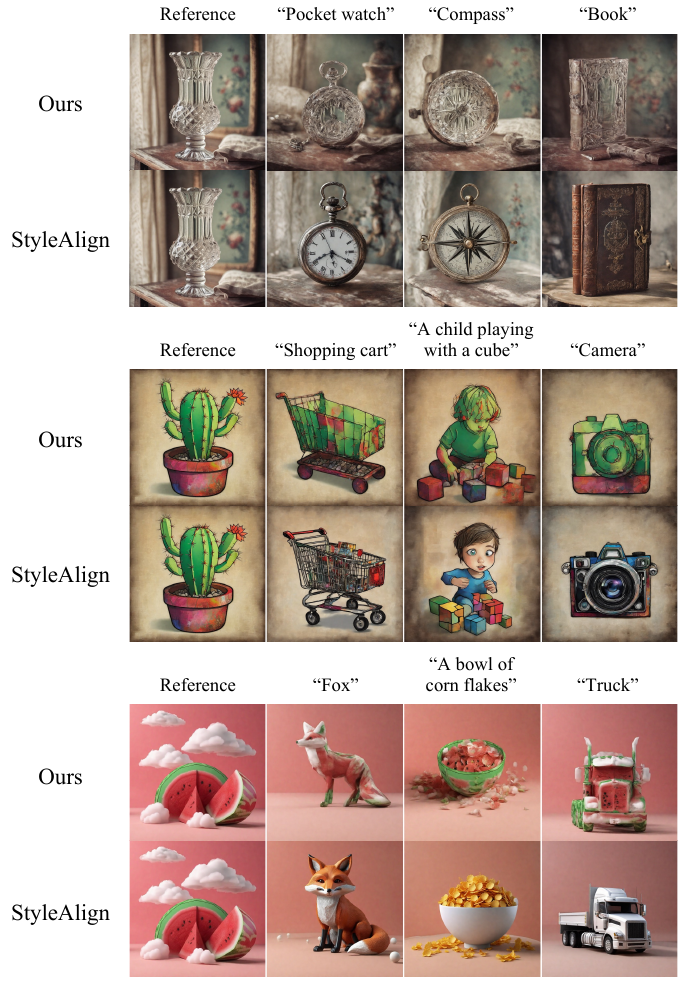}
        \caption{Definition of style is different between ours and StyleAlign.}
        \label{afig:vs_SA_definition_of_style}
    \end{minipage}

\end{figure}

\label{asec:selective}

\subsection{Style-Content prompt list}
\label{asec:style-content-prompt-list}
1. the great wave off kanagawa in style of Hokusai: (1) book (2) cup (3) tree 
 
2. fire photography, realistic, black background: (1) a dragon (2) a ghost mask (3) a bird 
 
3. A house in stickers style.: (1) A temple (2) A dog (3) A lion 
 
4. The persistence of memory in style of Salvador Dali: (1) table (2) ball (3) flower 
 
5. pop Art style of A compass . bright colors, bold outlines, popular culture themes, ironic or kitsch: (1) A violin (2) A palm tree (3) A koala 
 
6. A compass rose in woodcut prints style.: (1) A cactus (2) A zebra (3) A blizzard 
 
7. A laptop in post-modern art style.: (1) A man playing soccer (2) A woman playing tennis (3) A rolling chair 
 
8. A horse in colorful chinese ink paintings style: (1) A dinosaur (2) A panda (3) A tiger 
 
9. A piano in abstract impressionism style.: (1) A villa (2) A snowboard (3) A rubber duck 
 
10. A teapot in mosaic art style.: (1) A kangaroo (2) A skyscraper (3) A lighthouse 
 
11. A robot in digital glitch arts style.: (1) A cupcake (2) A woman playing basketball (3) A sunflower 
 
12. A football helmet in street art graffiti style.: (1) A playmobil (2) A truck (3) A watch 
 
13. Teapot in cartoon line drawings style.: (1) Dragon toy (2) Skateboard (3) Storm cloud 
 
14. A flower in melting golden 3D renderings style and black background: (1) A piano (2) A butterfly (3) A guitar 
 
15. Slices of watermelon and clouds in the background in 3D renderings style.: (1) A fox (2) A bowl with cornflakes (3) A model of a truck 
 
16. pointillism style of A cat . composed entirely of small, distinct dots of color, vibrant, highly detailed: (1) A lighthouse (2) A hot air balloon (3) A cityscape 
 
17. the garden of earthly delights in style of Hieronymus Bosch: (1) key (2) ball (3) chair 
 
18. Photography of a Cloud in the sky, realistic: (1) a bird (2) a castle (3) a ship 
 
19. A mushroom in glowing style.: (1) An Elf (2) A dragon (3) A dwarf 
 
20. The scream in Edvard Munch style: (1) A rabbit (2) a horse (3) a giraffe 
 
21. the girl with a pearl earring in style of Johannes Vermeer: (1) door (2) pen (3) boat 
 
22. A wild flower in bokeh photography style.: (1) A ladybug (2) An igloo in antarctica (3) A person running 
 
23. low-poly style of A car . low-poly game art, polygon mesh, jagged, blocky, wireframe edges, centered composition: (1) A tank (2) A sofa (3) A ship 
 
24. Kite surfing in fluid arts style.: (1) A pizza (2) A child doing homework (3) A person doing yoga 
 
25. anime artwork of cat . anime style, key visual, vibrant, studio anime, highly detailed: (1) A lion (2) A chimpanzee (3) A penguin 
 
26. A cactus in mixed media arts style.: (1) A shopping cart (2) A child playing with cubes (3) A camera 
 
27. the kiss in style of Gustav Klimt: (1) shoe (2) cup (3) hat 
 
28. Horseshoe in vector illustrations style.: (1) Vintage typewriter (2) Snail (3) Tornado 
 
29. play-doh style of A dog . sculpture, clay art, centered composition, Claymation: (1) a deer (2) a cat (3) an wolf 
 
30. A cute puppet in neo-futurism style.: (1) A glass of beer (2) A violin (3) A child playing with a kite 
 
31. the birth of venus in style of Sandro Botticelli: (1) lamp (2) spoon (3) flower 
 
32. line art drawing of an owl . professional, sleek, modern, minimalist, graphic, line art, vector graphics: (1) a Cheetah (2) a moose (3) a whale 
 
33. origami style of Microscope . paper art, pleated paper, folded, origami art, pleats, cut and fold, centered composition: (1) Giraffe (2) Laptop (3) Rainbow 
 
34. A crystal vase in vintage still life photography style.: (1) A pocket watch (2) A compass (3) A leather-bound journal 
 
35. The Starry Night, Van Gogh: (1) A fish (2) A cow (3) A pig 
 
36. A village in line drawings style.: (1) A building (2) A child running in the park (3) A racing car 
 
37. A diver in celestial artworks style.: (1) Bowl of fruits (2) An astronaut (3) A carousel 
 
38. A frisbee in abstract cubism style.: (1) A monkey (2) A snake (3) Skates 
 
39. Flowers in watercolor paintings style.: (1) Golden Gate bridge (2) A chair (3) Trees (4) An airplane 
 
40. A horse in medieval fantasy illustrations style.: (1) A castle (2) A cow (3) An old phone 

%/mnt/image-net-full/jaeseok/ip_adapter/ready_prompt_noise.ipynb
%%%%%%%%%%%%%%%%%%%%%%%%%%%%%%%%%%%%%%%%%%%%%%%%%%%%%%%%%%%%%%%%%%%%%%%%%%%%%%%
%%%%%%%%%%%%%%%%%%%%%%%%%%%%%%%%%%%%%%%%%%%%%%%%%%%%%%%%%%%%%%%%%%%%%%%%%%%%%%%

\end{document}